\documentclass[11pt]{scaleai-paper}
\usepackage{amsmath}
\usepackage{amsfonts}
\usepackage{amssymb}
\usepackage{booktabs}
\usepackage[square,numbers,sort&compress]{natbib}
\usepackage{xcolor}
\usepackage{hyperref}
\usepackage{todonotes}
\usepackage[most]{tcolorbox}

\definecolor{brandEvergreen}{HTML}{193A29}
\definecolor{brandForest}{HTML}{62A16E}
\definecolor{brandSage}{HTML}{9AB896}
\definecolor{brandSlate}{HTML}{839CB2}
\definecolor{brandNavy}{HTML}{273252}
\providecolor{scaleInk}{HTML}{212121}
\providecolor{scaleGray}{HTML}{929292}
\providecolor{scaleMediumGray}{HTML}{C7C7C7}
\providecolor{scaleLightGray}{HTML}{EAEAEA}
\providecolor{scalePanel}{HTML}{F2F2F2}

\tcbset{cardpanel/.style={enhanced, frame hidden, colback=scalePanel,
  arc=0pt, boxrule=0pt, left=7pt, right=7pt, top=5pt, bottom=5pt,
  before skip=7pt, after skip=7pt}}

\newtcolorbox{rubriccard}[2]{%
  cardpanel, unbreakable,  %
  before upper={\noindent\textcolor{brandEvergreen}{\small\sffamily\bfseries #1}%
    \hfill{\small\sffamily\bfseries\textcolor{scaleInk}{#2}}\par
    \vspace{2pt}{\color{scaleMediumGray}\hrule height 0.5pt}\vspace{4pt}},
}
\newcommand{\meta}[1]{#1}
\newcommand{\metasep}{, }

\newcommand{\rubline}[2]{%
  \par\addvspace{2pt}\noindent\hangindent=4.9em\hangafter=1
  \makebox[4.9em][l]{\textcolor{brandEvergreen}{\scriptsize\sffamily\bfseries\MakeUppercase{#1}}}#2\par}

\newtcolorbox{packetquote}[1]{%
  cardpanel, breakable,
  fontupper=\footnotesize,
  before upper={\textcolor{brandEvergreen}{\scriptsize\sffamily\bfseries\MakeUppercase{#1}}\par\vspace{3pt}},
}

\newcommand{\Nscenarios}{ten}
\newcommand{\Nitems}{1{,}257}
\newcommand{\ItemsMin}{74}
\newcommand{\ItemsMax}{170}
\newcommand{\Nbriefs}{1{,}020}

\newcommand{\colhead}[1]{\textcolor{brandEvergreen}{\scriptsize\sffamily\bfseries\MakeUppercase{#1}}}

\newenvironment{severitytable}{%
  \begin{center}\small
  \begin{tabular}{@{}l c l@{}}
  \textcolor{brandEvergreen}{\scriptsize\sffamily\bfseries SEVERITY}
  & \textcolor{brandEvergreen}{\scriptsize\sffamily\bfseries WEIGHT}
  & \textcolor{brandEvergreen}{\scriptsize\sffamily\bfseries MEANING} \\[1pt]}
  {\end{tabular}\end{center}}
\newcommand{\sevrow}[3]{#1 & #2 & #3 \\}

\newtcolorbox{figplaceholder}[1][4.5cm]{%
  enhanced, frame hidden,
  colback=white,
  borderline={0.6pt}{0pt}{scaleMediumGray,dashed},
  arc=0pt, boxrule=0pt,
  left=12pt, right=12pt, top=8pt, bottom=8pt,
  height=#1, valign=center,
  fontupper=\footnotesize\color{scaleGray},
  before upper={\textcolor{scaleGray}{\scriptsize\sffamily\bfseries FIGURE PLACEHOLDER}\par\vspace{4pt}},
}

\usepackage{listings}
\usepackage{placeins}  %
\IfFileExists{inconsolata.sty}{\usepackage[varqu,varl]{inconsolata}}{}
\newcommand{\promptlisting}[1]{%
  \tcbinputlisting{enhanced, breakable, listing only,
    colback=white, colframe=scaleMediumGray, boxrule=0.5pt, arc=3pt,
    left=8pt, right=8pt, top=6pt, bottom=6pt,
    before skip=8pt, after skip=8pt,
    listing options={extendedchars=true, literate={—}{{---}}3 {–}{{--}}2},
    listing file={#1}}}

\makeatletter
\providecommand{\@LN}[2]{}
\providecommand{\@LN@col}[1]{}
\makeatother

\newif\iflimitsinbody
\limitsinbodytrue

\makeatletter
\newcommand{\usecompactfigures}{%
  \NewCommandCopy\cmp@includegraphics\includegraphics
  \RenewDocumentCommand\includegraphics{s O{} m}{%
    \filename@parse{##3}%
    \IfFileExists{\filename@area\filename@base _acl.pdf}%
      {\cmp@includegraphics[width=\linewidth]{\filename@area\filename@base _acl.pdf}}%
      {\IfBooleanTF{##1}{\cmp@includegraphics*[##2]{##3}}{\cmp@includegraphics[##2]{##3}}}}}
\makeatother

\newsavebox{\fitbox}
\newcommand{\fittable}[1]{\sbox{\fitbox}{#1}\ifdim\wd\fitbox>\linewidth\resizebox{\linewidth}{!}{\usebox{\fitbox}}\else\usebox{\fitbox}\fi}

\usecompactfigures  %

\title{{\fontsize{20}{25}\selectfont MA-Bench: Measuring Operational Planning\\ in Language Models}}
\papertype{Preprint}
\contact{\texttt{rodney.lafuente@scale.com} \quad \texttt{robert.sincero@scale.com} \quad
\texttt{philsage@datamachines.com} \quad \texttt{yuan.xue@scale.com}\\[3pt]
{\normalsize\color[HTML]{1F5FBF}\hypersetup{pdfborder={0 0 0}}Data: \href{https://huggingface.co/datasets/ScaleAI/MA-Bench}{\underline{huggingface.co/datasets/ScaleAI/MA-Bench}}}}
\author[1]{Rodney Lafuente-Mercado}
\author[1]{Robert Sincero}
\author[2]{Phil Sage}
\author[1]{Yuan (Emily) Xue}
\affil[1]{Scale AI}
\affil[2]{Data Machines}

\begin{document}
\maketitle

\begin{abstract}
How well can a language model turn a large, incomplete and partly
contradictory body of documents into a plan when no single correct answer
exists? We introduce Mission Analysis (MA)-Bench, built on mission analysis, the first step
of military operational planning, in which a staff reads the full situation
and states the problem before anyone proposes a solution. Its 10 expert-built scenarios pair synthetic packets of ten to twelve documents with 1,257 expert-written pass/fail requirements, each containing information on their source, question type, and importance level. A reasoning-model judge grades each brief against the
requirements without seeing the documents; it grades like a fifth credentialed planner, leaving a
four-planner panel's agreement unchanged (Fleiss $\kappa$ 0.42 to 0.41), and a judge from a second provider reproduces the model ranking (Spearman 0.94). Across twenty frontier and open-weights models
the best gets 62\% of the rubric items right. Models are strongest on facts stated
in the documents (58\%) and weakest on judgments that must be inferred
(33\%), and six failures in ten are omissions rather than errors. MA-Bench
grades the determinations, not the prose. We release the packets, rubrics, prompts,
scorer and every graded brief with item-level verdicts.

\end{abstract}

\section{Introduction}

Large language models are increasingly used not only to retrieve information or answer questions, but to support complex planning and decision-making. However, evaluating such systems remains difficult. Most agent benchmarks measure whether a task with an externally verifiable outcome was completed: a software issue resolved, a website navigated, a set of explicit constraints met. Operational planning has no such outcome, and although expert rubrics now grade open-ended work in medicine and other professions~\citep{arora2025healthbench, akyurek2025prbench}, none covers operational planning. A plan can be fluent, plausible, and internally coherent while omitting a critical determination, overlooking conflicting evidence, hardening an uncertain assessment into fact, or violating a governing constraint. Nothing in such a brief is wrong in a way a reference answer would catch. What is wrong is what is missing. Measuring that requires knowing, for a given situation, the full set of determinations a competent staff would make and checking each one. We study this problem through mission analysis, the step of the Joint Planning Process in which a staff transforms higher direction and a large body of operational information into a defined problem before developing courses of action. The capability targeted is neither retrieval nor fluency, but judgment under evidence: which determinations the situation demands, and whether the system makes them.

\begin{figure}[t]
\centering
\includegraphics[width=\textwidth]{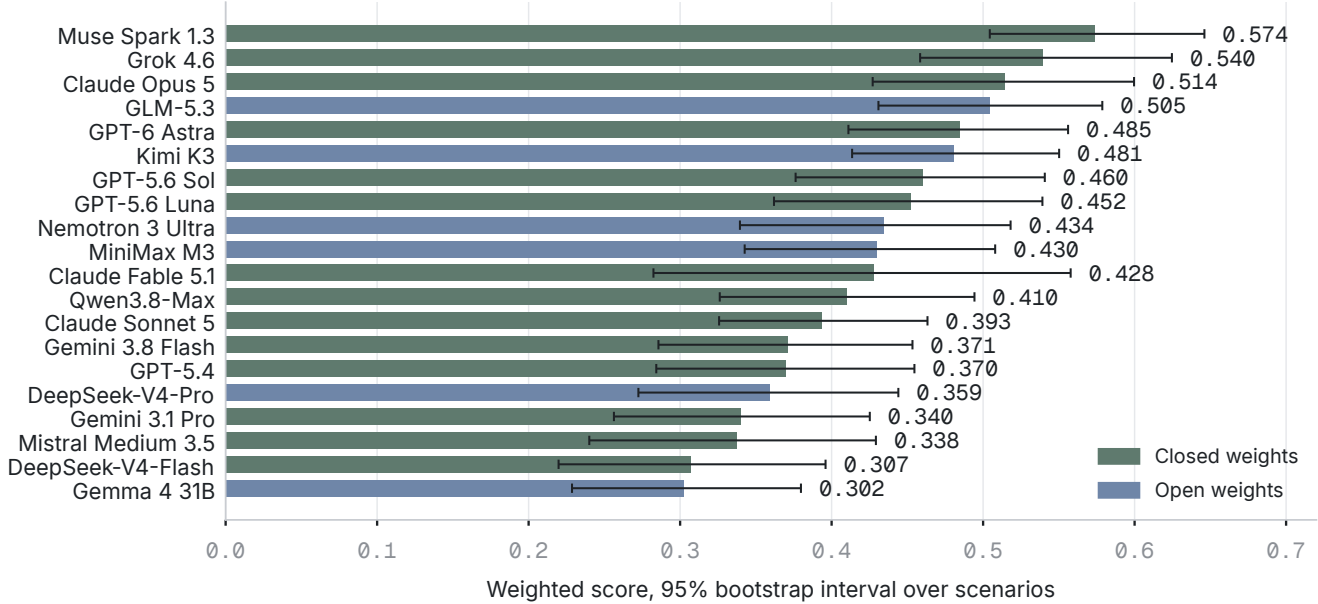}
\caption{\textbf{Weighted score by model, production judge.} Benchmark
condition (3,000-word cap, medium effort), all twenty systems; whiskers are 95\% bootstrap intervals over the ten scenarios, and a perfect
brief scores 1.}
\label{fig:main}
\end{figure}

\begin{figure*}[t]
    \centering
    \includegraphics[width=\textwidth]{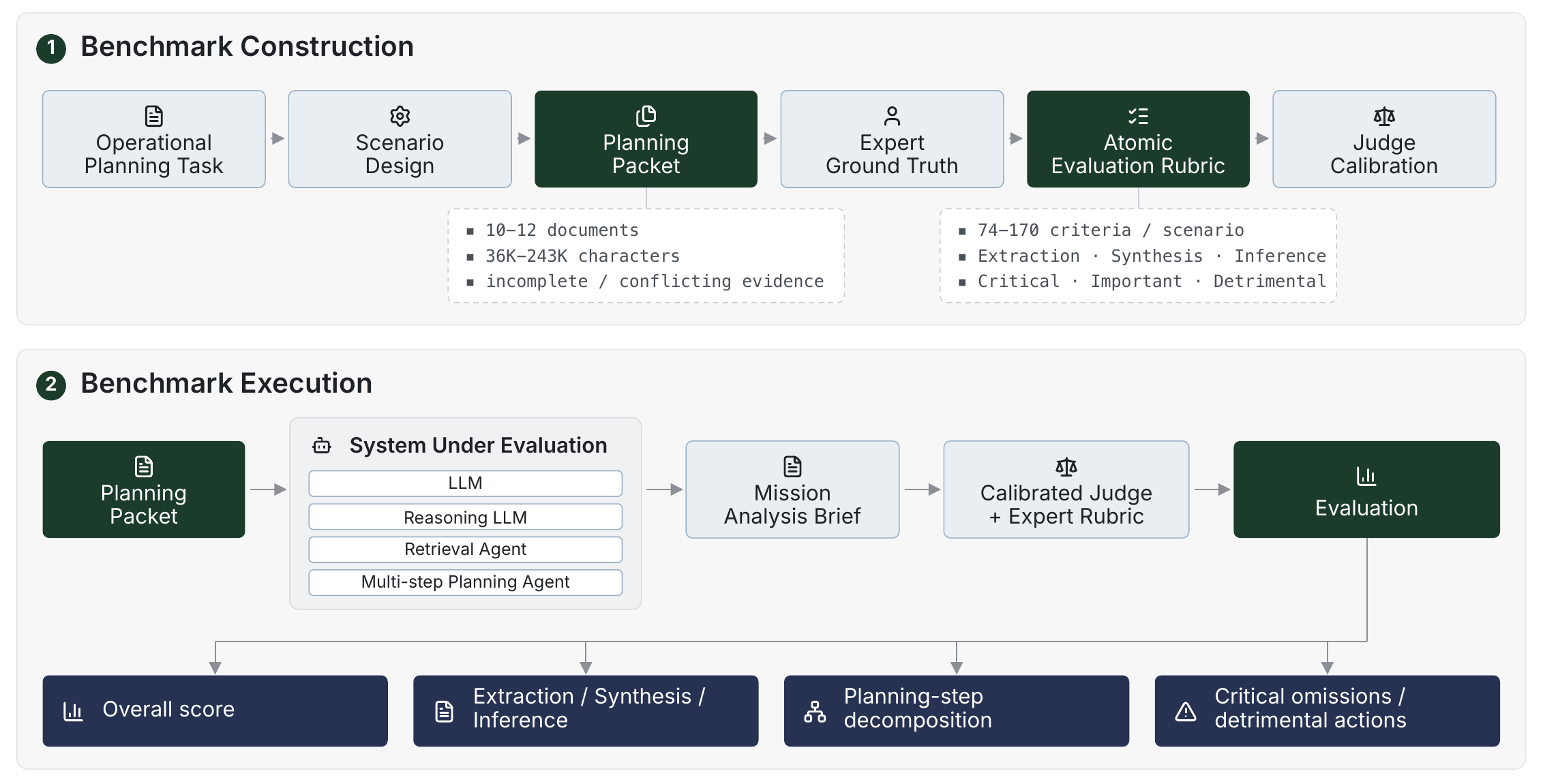}
    \caption{\textbf{MA-Bench construction and evaluation pipeline.}
    Top: benchmark construction, from operational planning task design through
    scenario construction, planning packets, expert ground truth, atomic rubrics,
    and judge calibration. Bottom: benchmark execution, where a planning packet is
    provided to the system under evaluation, the resulting mission analysis brief
    is graded using the calibrated judge and expert rubric, and performance is
    decomposed by overall score, reasoning level, planning step, and critical or
    detrimental errors.}
    \label{fig:mabench_pipeline}
\end{figure*}

We introduce MA-Bench, a benchmark for measuring this capability. Each of its \Nscenarios{} scenarios provides a synthetic planning packet written and validated by planning experts and asks for a mission analysis brief. Instead of comparing the brief with a reference text, MA-Bench grades it against an expert-authored set of atomic requirements grounded in doctrine and scenario evidence, \Nitems{} in all. Items distinguish information that can be extracted directly, conclusions that require synthesis across evidence, and planning judgments that require inference; they also distinguish omissions from detrimental actions, so the benchmark captures failures that accuracy-style metrics obscure. The protocol grades the brief and is agnostic to how it was produced, so the system under test can be a language model given the packet in a single prompt or a multi-step planning system. This paper evaluates twenty models under one fixed protocol (Figure~\ref{fig:main}).

Our main contributions are the evaluation methodology and the dataset, and the evaluation of twenty models shows what they measure:
\begin{itemize}
 \item A formulation of operational planning evaluation for open-ended, high-stakes planning tasks with heterogeneous evidence and no unique reference answer.
 \item MA-Bench, consisting of \Nscenarios{} expert-validated scenarios, long multi-document planning packets, and \Nitems{} expert-authored atomic evaluation items, with a structured evaluation methodology separating extraction, synthesis, and inference while explicitly scoring both omissions and detrimental actions.
 \item An empirical evaluation of twenty frontier and open-weights models with a judge whose rubric grading is checked against credentialed planners, analyzing overall performance, capability decomposition, failure modes, and evaluation reliability.
\end{itemize}

\section{Mission Analysis as an Evaluation Problem}
\label{sec:problem}

Operational planning turns direction from a higher headquarters into a plan
that forces can execute. In the United States joint doctrine, it follows the Joint
Planning Process~\citep{jp50}, and mission analysis is the first step in which
the staff produces   its own judgments. Given the higher headquarters' order, the intelligence picture, the forces
available, and the commander's initial guidance, the staff must state the
problem before anyone proposes a solution: the tasks that are specified,
implied and essential, the constraints and restraints it must operate within,
the facts it can rely on and the assumptions it must make to keep planning,
the information the commander still needs before deciding, the risks to the
mission and to the force, and a restated mission in a single sentence. These
products are the foundation of every later step, so a determination missed or
a fact hardened here propagates through the plan.

For researchers, this is a long-context reasoning task with expert ground
truth and no reference answer over a large, heterogeneous, incomplete, and
partly conflicting packet; for practitioners, it is a measured instrument for a
task language models are already being asked to perform.

Formally, let $X = \{d_1, \ldots, d_n\}$ denote a planning packet of $n$
source documents and $A$ the planning system under evaluation, which produces
an analytical artifact $y = A(X)$. Operational planning does not admit unique
reference output: two briefs may differ in wording, organization, and emphasis
while both are operationally sound, so the objective of the evaluation is not
similar to a reference, but whether $y$ makes the determinations that sound
planning requires, represented as a set $R = \{r_1, \ldots, r_m\}$ of atomic
requirements, each specifying the determination, the evidence or doctrinal
basis that supports it, the reasoning needed to reach it, and the consequence
of omitting it or introducing an incorrect or detrimental conclusion. The
benchmark asks whether $y$ satisfies each $r_i$, independently of prose,
formatting, and organization. Failures are not limited to missing
information: an artifact can introduce unsupported conclusions, overstate
uncertain evidence, or violate a governing constraint, so we distinguish
positive requirements, determinations a sound artifact should contain, from
negative requirements, specific detrimental moves it should not make, and
score both. Two artifacts of similar completeness can differ sharply in
quality if one merely omits a useful determination while the other
confidently states a materially wrong one.

\section{Related Work}
\label{sec:related}

MA-Bench sits at the intersection of five literatures, reviewed in
Appendix~\ref{app:related}: agent and occupation-scale benchmarks with verifiable or
expert-graded outcomes~\citep{jimenez2024swebench, mialon2024gaia,
zhou2024webarena, xie2024osworld, yao2024taubench, patwardhan2025gdpval,
mazeika2025rli}, planning evaluations with a checkable end
state~\citep{valmeekam2023planbench, xie2024travelplanner,
zheng2024naturalplan}, model-based grading and its position, length and
self-preference biases~\citep{zheng2023mtbench, liu2023geval,
wang2024fairevaluators, dubois2024lcalpaca, panickssery2024selfpreference},
expert-authored atomic rubrics~\citep{arora2025healthbench,
akyurek2025prbench, wang2025profbench, gunjal2025rubrics}, and language models in military
decision-making~\citep{rivera2024escalation, lamparth2024wargames,
geist2024limits}. None grades a doctrinal staff product whose value lies
in its defensibility, and JP~5-0~\citep{jp50} supplies the standard MA-Bench
grades against.

\section{Benchmark Design}
\label{sec:benchmark}

Figure~\ref{fig:mabench_pipeline} gives an overview. Each MA-Bench scenario has
three fixed components: an input packet modeled on the
one a planning staff would receive, an expert-authored rubric, and a grading
protocol under a calibrated model judge.

\subsection{Scenario packets}
\label{sec:packets}

Each scenario is a synthetic, expert-validated operational crisis with real
geography and real actors, written in the format of an actual staff packet
(strategic context, intelligence summary and preparation brief, orders of
battle, country study and geography, the initiating order, legal annexes, the
commander's initial planning guidance and running staff estimates), 36k to
243k characters across ten to twelve documents. No packet or rubric text
exists on the public internet, so exact-text contamination is ruled out by
construction; models have seen the doctrine, staff templates, and real-world
context the scenarios draw on, and the benchmark intends that, since doctrine
is the standard a brief is held to,  knowing it does not supply what a
particular packet says or implies (Section~\ref{sec:limits}).

The material is staff prose that describes a situation while leaving key
judgments implicit. The passage below moves from the collapse of the United
Nations route through the no-fly zone declaration to regime-change planning
that must stay separate from the protective operation, and a brief earns
credit by carrying those judgments forward, not by restating the events:

\begin{packetquote}{Gulf Guardian\metasep Strategic Context, paras.~4 and 7}
The President sought a United Nations Security Council resolution authorizing a
force to protect Cuban civilians. The Russian Federation and the People's
Republic of China vetoed it, and both now describe the atrocity reporting as
imperialist propaganda intended to manufacture a pretext for invasion. Acting
without a Security Council mandate, the President declared a no-fly zone over
the Cuban coast and stated that any further killings would be dealt with
forcefully. The United States thus acts on a civilian-protection rationale
under customary international law, over the objection of two great powers, and
the legitimacy of that case is now itself a strategic center of gravity.

\vspace{4pt}{\color{scaleGray}\itshape\footnotesize [\,\ldots\,]}\vspace{2pt}

The direction to prepare, on order, plans for operations ashore and for regime
change hangs over the whole effort. It is planning only, gated on a decision by
the National Command Authority, and it must be held apart, in fact and in
perception, from the protective operation that begins now, lest it hand Havana,
Moscow, and Beijing the invasion narrative they are seeking.
\end{packetquote}

The corpus is balanced across five design axes (plan type, domain,
stakeholder mix, time pressure, and operational-environment complexity; target
distribution in Appendix~\ref{app:corpus}). A source index records each
document's visibility and the rationale when it is withheld: withheld
documents are downstream products of a completed mission analysis (component
orders, the commander's critical information requirements, event templates)
that ground the rubric during authoring and never reach the system under
test. Exclusion is enforced by construction and checked before every run: the
released document set holds only the visible documents, each with a content
hash in the bundle manifest, the packet and any multi-step system's retrieval
index are assembled from that set alone, and the grading runner verifies
every manifest hash before its first call and writes the artifact hashes to
its run log, which ships with the release, so each reported number is tied to
a specific packet build. The packets are deliberately
incomplete: some reporting is unconfirmed and marked as such, and some
figures from different originators do not reconcile. Recognizing these limits
is itself scored.

\subsection{Rubrics}
\label{sec:rubrics}

Each scenario has one expert-authored rubric of \ItemsMin{} to \ItemsMax{}
items, \Nitems{} in all, authored against the packet, reviewed by
credentialed planners, versioned and frozen before evaluation, so the same
version grades every system in a comparison. Each item meets six
requirements: atomic (one claim or behavior), binary, observable in the
submitted brief rather than in intent or process, grounded in doctrine or
packet evidence with a resolvable citation, self-contained (gradable without
opening the packet), and non-duplicative. A directive to ``protect
infrastructure and deter sabotage'' becomes two items, and vague completeness
tests, prose quality, and anything satisfiable by copying obvious text are
prohibited.

The \textbf{tier} records how the answer is grounded. Tier~1 items are stated
outright in a single document (Appendix~\ref{app:cards} shows one). Tier~2
items require synthesis across several facts. Tier~3 items are planning
judgments that appear in no document and must be inferred; they are the most
valuable and the hardest. A Tier~3 item:

\begin{rubriccard}{C5:GG-S1-05v1}{\meta{Tier 3}\metasep\meta{I}\metasep\meta{Step S1}}
\small
\rubline{Question}{Does the response surface the tension between protecting
civilians now and the still-gated regime-change tasker?}
\rubline{Pass}{Frames the tension and the need to keep the two apart in fact
and in perception.}
\rubline{Fail}{Ignores the tension or merges regime change into the base
operation.}
\rubline{Source}{Commander's Initial Planning Guidance \S 2.}
\end{rubriccard}

The \textbf{severity} is a signed importance label fixed at authoring time;
the sign says whether the item is desirable and the magnitude how much it
moves the score, under a weight mapping declared and frozen before each run
(Section~\ref{sec:scoring}):

\begin{severitytable}
\sevrow{Critically important (CI)}{$+9$}{missing it breaks the analysis}
\sevrow{Important (I)}{$+5$}{missing it materially weakens the analysis}
\sevrow{Slightly important (SI)}{$+2$}{useful nuance}
\sevrow{Detrimental (D)}{$-5$}{a recoverable wrong move; penalized when present}
\sevrow{Critically detrimental (CD)}{$-9$}{a wrong move partial credit must not offset}
\end{severitytable}

Negative items name a specific wrong move, such as treating a
moderate-confidence attribution as confirmed fact, and the brief is penalized
when the move appears (Appendix~\ref{app:cards}).

Each item is also assigned to one of the twelve mission analysis steps
(higher mission and intent, operational environment, task analysis,
limitations, information requirements, assumptions, risk, restated mission,
decision points, facts, forces and objectives), so scores decompose by step.
Ordering, formatting, and style earn nothing.

\subsection{Grading and scoring}
\label{sec:judge}
\label{sec:scoring}

A single reasoning model grades each brief against the full rubric in one
call, returning only the items the brief fails, each with quoted evidence and
a rationale tied to the fail condition, in a schema-enforced format; unlisted items pass. For a negative item, the
pass condition is the presence of the flagged wrong move, so a listed
negative item is a mistake the brief avoided and an unlisted one a mistake it
committed. Malformed responses are rejected and retried, and no accepted
verdict lists fewer than 13 items. An item the brief does not clearly satisfy
fails, with no partial credit, and the judge never sees the packet, so grading measures the brief against the rubric
rather than the judge's reading of the scenario. The judge is selected by
measurement: candidates are compared on the expert-graded sample of
Section~\ref{sec:irr} and admitted only if their chance-corrected agreement
with the expert majority is at least the experts' mean agreement with one
another and their $F_1$ on the fail class reaches $0.80$. The production
judge clears both gates, $\kappa = 0.54$ against an expert mean of $0.43$ and
$F_1 = 0.82$ on 190 item verdicts graded by four planners; the judge was frozen before
any benchmark brief was graded, and every result is reproduced under a second
judge from another provider (Section~\ref{sec:judge_sens}). Between the
expert study and the final freeze one item was restored, five were reworded
and thirteen negative items were corrected for polarity
(Section~\ref{sec:irr}).

The base score is the unweighted fraction of rubric items passed. The
severity-weighted score uses the declared mapping: with $P$ the positive items
(CI, I, SI), $N$ the negative items (D, CD), and $w_i$ the weight of item $i$,
\begin{equation*}
S \;=\; \frac{\sum_{i \in P} w_i\,[\text{$i$ satisfied}]\;-\;\sum_{j \in N} |w_j|\,[\text{$j$ present}]}{\sum_{i \in P} w_i},
\end{equation*}
so a perfect brief scores $1$, a brief that meets no requirement and makes no listed wrong move scores $0$, and wrong moves are charged in the same units
as omissions. Scores therefore place systems on one scale for comparison and are not calibrated against human-written briefs, which the benchmark does not include (Section~\ref{sec:limits}). The score is not clamped, since flooring at zero would hide the
behavior  negative items are meant to catch (worked example in
Appendix~\ref{app:scoring}); seeds are averaged within a scenario before any
comparison, and the benchmark is agnostic about how a brief was produced.

\section{Experiments}
\label{sec:results}

\subsection{Setup}
\label{sec:setup}

Sixteen models are evaluated in the benchmark condition: Claude Opus 5, Claude
Fable 5.1, GPT-6 Astra, GPT-5.6 Sol and luna, Gemini 3.1 Pro, Grok 4.6, Mistral Medium 3.5, Muse
Spark 1.3 and Qwen3.8-Max through their API providers, and the
open-weights GLM-5.3, DeepSeek-V4-Pro, Kimi K3, MiniMax M3, Nemotron 3 Ultra
and Gemma 4 31B; Claude Sonnet 5, GPT-5.4, Gemini 3.8 Flash and
DeepSeek-V4-Flash form a capability ladder. Every model receives the same task
prompt naming the twelve mission analysis steps, followed by the full packet
(Appendix~\ref{app:prompts}), a 3,000-word cap enforced before grading,
medium reasoning effort where we set it and the provider default otherwise
(Appendix~\ref{app:gen} lists each run's setting), and at least three
independent briefs per scenario, five for the three models run first; outputs
that end in a provider content filter are graded as delivered and counted as
refusals; the benchmark comprises \Nbriefs{} graded briefs across these
conditions, and 540 uncapped briefs extend the length ablation to every model (Appendix~\ref{app:ablation}). The cap is part of the protocol. Models are told the limit in the prompt, it is about the length of a staff mission analysis brief, and it keeps scores from tracking length, which they follow far more closely without it (Spearman 0.70 against 0.26), while the ranking largely holds (Spearman 0.95, Appendix~\ref{app:ablation}). The production
judge is gpt-5.6 at high reasoning effort with the frozen prompt of
Section~\ref{sec:judge}, one call per brief and no access to the packet;
Gemini 3.1 Pro regrades every brief, and a common subset of 216
benchmark-condition briefs (six scenarios, ten models, every seed) is
regraded by both Gemini 3.1 Pro and Claude Opus 5
(Section~\ref{sec:judge_sens}). The scenario is the unit of analysis: seeds
are averaged within a scenario, means and 95\% bootstrap percentile intervals
are taken over the \Nscenarios{} scenarios~\citep{efron1993bootstrap},
comparisons between conditions of one model are paired by scenario, and we
report the unclamped weighted score of Section~\ref{sec:scoring}, items
correct (a negative item counts as correct when its mistake is absent), and
the positive pass rate.

\subsection{Main results}
\label{sec:main}

Figure~\ref{fig:main} reports the benchmark condition, with the numbers in
Table~\ref{tab:main} (Appendix~\ref{app:main}). The best model, Muse Spark 1.3, gets 62\% of rubric items right
for a weighted score of 0.574, with Grok 4.6 at 0.540, Claude Opus 5 at 0.514,
and GLM-5.3, the best open-weights model, at 0.505. The sixteen models span
0.30 to 0.57, so the benchmark separates systems that are close on other
leaderboards. No model approaches the ceiling, and with
intervals about $\pm 0.08$ wide, adjacent models are not separated; only the
ends of the table are. Between two fifths and two thirds of the determinations a planning staff would require are missing from every
brief, and every brief commits three to five flagged wrong moves, about two of
them critical.

Three features of the table recur throughout the paper. GPT-6 Astra commits the
fewest wrong moves (3.1 per brief, 1.4 critical) while ranking fifth on coverage, so coverage and commission are separable qualities that one
accuracy figure would conflate. The two GPT-5.6 variants land within 0.01 of
one another, the resolution expected from two serving configurations of one
model. Claude Fable 5.1's wide interval is a refusal effect: its content
filter terminated every brief on Jagged Anvil and four of five on Pacific
Harrier, mid-sentence and without a refusal message, and those outputs are
graded as delivered, because a planning system that declines two crises in ten is a fact about
the system. Without those nine briefs, it scores 0.510 (0.432, 0.595) over the
nine scenarios that keep a brief, level with Claude Opus 5.

\begin{figure}[t]
\centering
\includegraphics[width=\textwidth]{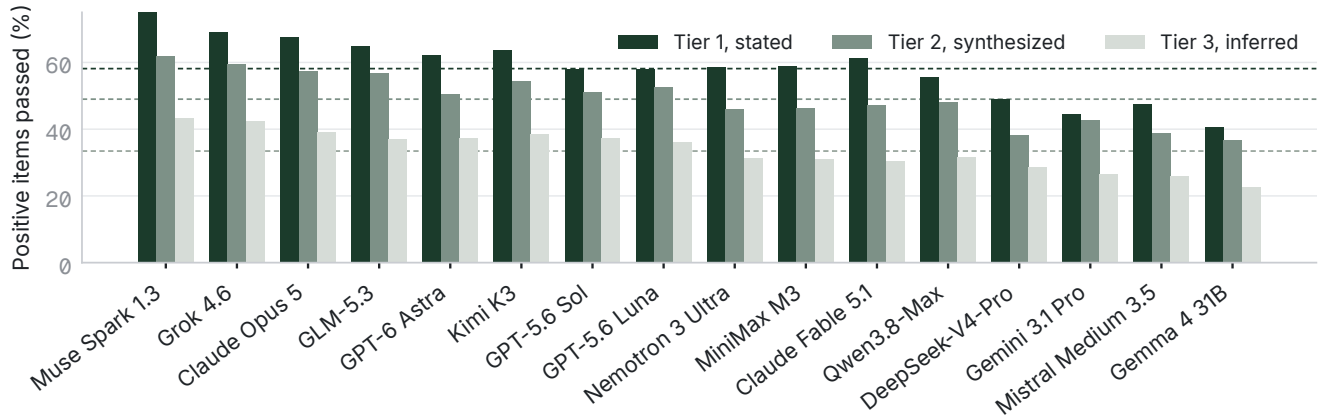}
\caption{\textbf{Pass rate by grounding tier and model}, benchmark condition,
production judge. Dashed lines are the pooled rates (58.1, 49.0 and 33.4).}
\label{fig:tiers}
\end{figure}

\subsection{Where models fail}
\label{sec:decomp}
\label{sec:failures}

Pooled over the benchmark condition, positive items pass at 58.1\% when the
answer is stated in a single document (Tier~1), 49.0\% when it must be
assembled from several (Tier~2) and 33.4\% when it must be inferred (Tier~3),
and the gradient holds for every model (Figure~\ref{fig:tiers}). Tier~3 items
make up 36\% of the rubric, so the planning judgments the benchmark was built to
measure  also account for the largest share of the missing credit.
Critically important items pass at 56.9\% and important items at
37.5\%:
models make the headline determinations and miss the second-order ones a
reviewing staff would consider material. By step (Figure~\ref{fig:steps},
Appendix~\ref{app:steps}), models are strongest where the packet states the
answer and the step carries it forward (constraints and restraints 72\%, higher mission and intent 66\%) and weakest on the steps that look forward
from the situation to the decisions it will force (commander's critical
information requirements 31\%, decision points and named areas of interest
38\%, force and resource analysis 41\%). The ordering holds on the 285 briefs that
never reached the cap (28\% and 36\% on the first two), so it is not an
artifact of truncation. Models describe the situation they are given far
better than they anticipate the decisions it will demand.

The judge's rationale for every failed item says what kind of failure it was.
A keyword classification
of the 31,882 failed positive items assigns 60.2\% to outright omission, 8.2\% to an incomplete
treatment, 2.5\% to content filed under the wrong step, and 1.6\% to incorrect or
unsupported content, with 28\% unassigned, and the proportions barely move across models, tiers, or
severities (Appendix~\ref{app:failures}). Models rarely say something the rubric can catch as wrong; they leave things
out, which is why a reference-similarity or accuracy metric
would miss most of what the rubric catches: there is no wrong sentence to
detect. Wrong moves are the exception the negative items exist to catch. A flagged mistake is committed on 25.9\% of detrimental
and 28.3\% of critically detrimental opportunities, about four per brief, more
often on plainly stated traps (35\% of Tier~1 opportunities against 21\% of Tier~3) such as treating
an unconfirmed report as fact, and concentrated on the known-facts step, where the flagged
mistake is committed on 44\% of opportunities. Only the positive items pay.

One item from the first of Claude Opus 5's briefs on Gulf Guardian shows what
these failures look like in practice: the item, what the brief did, and the
judge's rationale verbatim.

\begin{rubriccard}{C5:C-06v1}{\meta{Tier 3}\metasep\meta{CI}\metasep\meta{Step S2}}
\small
\rubline{Item}{Does the analysis describe how the commander's operational
approach of protecting civilians will weaken the Cuban center of gravity?}
\rubline{Brief}{Lists the regime's critical vulnerabilities in one section and
calls for the release of evidence of the killings in another.}
\rubline{Judge}{``The response lists regime vulnerabilities and separately
calls for evidence release, but never explains how civilian protection
directly weakens coercive control and the regime's blame-shifting
narrative.''}
\end{rubriccard}

This is the characteristic Tier~3 failure. Every fact the inference needs is
on the page, and the inference is not. The characteristic Tier~1 failure, from
the same brief, is traced in full in Appendix~\ref{app:cards}: the packet
names seven partner governments in one table and the brief, which cites that
table three times, writes ``willing regional partners''. Nothing in either
would trouble a reader judging fluency. Both would trouble the staff that had
to plan from the brief. The appendix adds a third card, from the
information-requirements step that scores lowest of the twelve.

\subsection{Length, reasoning effort and model scale}
\label{sec:ablations}

Scores are not a proxy for length. Within the capped condition, the rank
correlation between a model's raw brief length and its score is 0.26 (Pearson
0.15) across the twenty systems, and the longest writer, Gemini 3.8 Flash,
ranks fourteenth. Removing the cap raises a model's score (GPT-6 Astra by 0.078,
GPT-5.4 by 0.123), so models hold more rubric-relevant content than a
staff-length brief can carry (Appendix~\ref{app:ablation}).

Neither more compute nor a bigger model closes the gap on its own. Moving
reasoning effort from low to high shifts a model's score by at most 0.08
across five models, with no consistent sign, even as reasoning tokens rise by
one to two orders of magnitude (Appendix~\ref{app:ablation}). Within a family,
the stronger member wins in three of four cases: Claude Opus 5 over Sonnet 5 by 0.12, GPT-6
Astra over GPT-5.4 by 0.115, and DeepSeek-V4-Pro over V4-Flash by 0.05, with the
Gemini pair tied.

\subsection{Sensitivity to the judge}
\label{sec:judge_sens}

\begin{table}[t]
\centering
\footnotesize
\setlength{\tabcolsep}{5pt}
\begin{tabular}{@{}l r r r l@{}}
\toprule
\colhead{Judge pair} & \colhead{Agreement} & \colhead{Cohen kappa} & \colhead{Spearman} & \colhead{Own-family excess} \\
\midrule
gpt-5.6 vs Gemini 3.1 Pro & 85.1\% & 0.701 & 0.92 & $+0.004$ / $+0.005$ \\
gpt-5.6 vs Claude Opus 5 & 80.0\% & 0.610 & 0.94 & $-0.016$ / $-0.029$ \\
Gemini 3.1 Pro vs Claude Opus 5 & 83.8\% & 0.675 & 0.89 & $-0.026$ / $-0.026$ \\
\bottomrule
\end{tabular}
\caption{\textbf{Agreement among three judges} on a common set of 216
benchmark-condition briefs (six scenarios, ten models, every seed, 27,540
item verdicts per pair): item agreement, Cohen kappa, Spearman correlation of
the ten model means, and each judge's self-preference excess on its own
family (first / second judge).}
\label{tab:judges}
\end{table}

The choice of judge could, in principle, determine every number above, so tjudges from two other providers regrade the briefs with the same frozen
prompt. Gemini 3.1 Pro regrades all 1,050 briefs and agrees with the production
judge on 84.0\% of 131,985 item verdicts, Cohen's $\kappa =
0.678$~\citep{cohen1960kappa}, with a Spearman correlation of 0.94 between
the two judges' means over the sixteen benchmark-condition models. On a common set of 216 briefs (six scenarios, ten
models, every seed) all three judges are compared pairwise
(Table~\ref{tab:judges}): agreement runs from 80.0\% to 85.1\% ($\kappa$ 0.61
to 0.70), in every case above the agreement between two human experts on this
task (Section~\ref{sec:irr}), and the rankings are stable (Spearman 0.89 to
0.94). The alternate judges are more lenient by a roughly constant offset,
which moves every model together and leaves the order intact
(Appendix~\ref{app:judges}). Self-preference, a documented failure mode of
model judges~\citep{panickssery2024selfpreference, zheng2023mtbench},
is measured rather than assumed: for each pair of judges we take the gap
between their scores on one judge's own model family and subtract the same
gap on all other families. Across the
three pairs the excess ranges from $-0.029$ to $+0.005$, so no judge favors its
own family by more than 0.03,
less than the width of any single model's interval.

\subsection{Agreement with experts}
\label{sec:irr}

\begin{narrowtable}[t]
\centering
\footnotesize
\setlength{\tabcolsep}{5pt}
\begin{tabular}{@{}l r r r r r@{}}
\toprule
\colhead{Severity} & \colhead{Items} & \colhead{Agree.} & \colhead{kappa} & \colhead{Expert fail} & \colhead{Judge fail} \\
\midrule
CI & 51 & 75\% & 0.49 & 53\% & 47\% \\
I  & 47 & 85\% & 0.65 & 66\% & 72\% \\
SI & 24 & 92\% & 0.83 & 62\% & 54\% \\
D  & 20 & 70\% & 0.37 & 75\% & 55\% \\
CD & 22 & 64\% & 0.21 & 91\% & 55\% \\
\midrule
All & 164 & 78\% & 0.54 & 66\% & 57\% \\
\bottomrule
\end{tabular}
\caption{\textbf{Judge against a four-expert panel}, by severity, on the 164
of 190 expert-graded verdicts with a majority (26 split two against two).
Fail columns are the shares each side failed. Candidate judges are compared
in Table~\ref{tab:candidates}.}
\label{tab:irr}
\end{narrowtable}

Four JPME-credentialed planners independently graded the same 190 item
verdicts, ten briefs by nineteen items covering every tier and severity, with
the information the judge receives and not the packet (Table~\ref{tab:irr}).
On 26 items the experts split two against two, so judge agreement is measured
on the 164 with a majority. The experts agree with one another at Fleiss
$\kappa = 0.42$~\citep{fleiss1971kappa}, pairwise 0.30 to 0.60 with mean
0.43, which bounds how well any judge can be validated against them; the production
judge agrees with the majority on 78.0\% ($\kappa = 0.54$, $F_1$ on the fail
class 0.82), and treated as a
fifth rater its mean pairwise $\kappa$ with them is 0.40, inside their range,
leaving the panel's Fleiss $\kappa$ unchanged (0.42 to 0.41). Holding out each planner in turn, the judge agrees with the majority of the other three at mean $\kappa = 0.46$, against 0.52 for the held-out planner (0.42 to 0.62), and a three-planner majority has no ties. Agreement is
flat across tiers, $\kappa$ 0.53, 0.56 and 0.51 on Tiers 1 to 3, so the judge
is no less reliable on the inference items the benchmark leans on. The prompt was
chosen on this sample by agreement alone: the stricter of two scoring-rule
variants won on a pre-declared half of the items and on all of them (0.61 and
0.54 against 0.52 and 0.46) and was frozen before any benchmark brief was
graded; a repeat run and the 137 items untouched by the final rubric revision
move $\kappa$ by 0.004 and to 0.51 (Appendix~\ref{app:judges}). The alternate
judges reach 0.43 (Gemini 3.1 Pro) and 0.34 (Claude Opus 5) against the same
majority, so only the production judge clears the gate of
Section~\ref{sec:judge}. Agreement is high on positive items and lower on
negative ones for experts and judge alike (Table~\ref{tab:irr}); the graders'
notes trace the drop to polarity, since a pass on a negative item means the
mistake is present, and thirteen items worded the other way round were
corrected in a documented errata before benchmark grading.

\section{Discussion and Conclusion}
\label{sec:limits}
\label{sec:conclusion}

MA-Bench targets a planning capability that no verifiable-outcome benchmark
reaches: turning a large, incomplete and partly contradictory body of evidence
into the determinations a staff needs before it can plan. The best current
model gets 62\% of the rubric items right; the missing share sits in the
inference tier and in the steps that look forward from the situation to the
decisions it will force, and the failures are omissions rather than errors,
invisible to any metric that checks what a brief says instead of what it
should have said. Three judges from three providers and four human experts
see the same picture, which we take as evidence that the rubric is applied
consistently and that the results do not depend on one grader. Every score is a model's
verdict: the judge grades like a fifth credentialed planner,
leaving the panel's agreement unchanged, two judges from other
providers reproduce the ranking at Spearman 0.89 and above with no self-preference beyond 0.03, and every verdict carries the quoted evidence
and rationale on which it rests. Planners graded model briefs in the expert study but wrote none of their
own, so the paper has no human baseline. Real planning packets are too
sensitive to use, and a human-written brief takes a planning team a week or
more, so we built expert-written synthetic packets and spent that expert time
on rubric items instead. The expert study shows that the judge applies the
rubric as planners do. Showing that a higher score means a better mission
analysis needs a holistic expert rating of briefs against the packets, and a
small set of human-written briefs would anchor the scale.

\iflimitsinbody  %
The remaining limits are properties of the instrument as run, each recorded
in the released verdicts. Agreement on detrimental items is lower for the
judge and the experts alike ($\kappa$ 0.21 to 0.37), and negative items make up
12\% of the rubric and enter the score subtractively. Bootstrap intervals over
ten scenarios are about $\pm 0.08$ wide, so adjacent models are not separated,
while the ends of the table are. No packet or rubric text exists outside the
benchmark, and models' knowledge of joint doctrine is intended: doctrine says
what a mission analysis should contain, not what this packet says or implies.
The benchmark covers one step of one planning process in one doctrinal
tradition; the length cap truncates many briefs, and the uncapped ablation
prices the cost, and provider content filters ended nine briefs, which are
counted.
\fi
Every packet, rubric, prompt, brief and verdict is released so that
the next model can be measured against the same standard.

\section*{Data availability}
The scenario packets, rubrics, scoring code, briefs and verdicts are released at
\url{https://huggingface.co/datasets/ScaleAI/MA-Bench}.

\clearpage  %
\bibliographystyle{abbrvnat}
\bibliography{references}

\AddToHook{cmd/appendix/after}{\noindent{\LARGE\bfseries Appendix}\par\bigskip}  %
\clearpage
\appendix

\FloatBarrier
\section{Extended related work}
\label{app:related}

MA-Bench sits at the intersection of five bodies of literature: agentic task
benchmarks, planning evaluation,  model-based (LLM-as-judge)
grading and its reliability, rubric-based evaluation of open-ended professional
work, and the emerging study of language models in military decision-making.
MA-Bench addresses a gap spanning all five: no public benchmark uses expert-authored atomic rubrics to grade a planning system's performance on the first step of the Joint Planning Process against codified documents.

\subsection{Agentic task benchmarks and the fluency bias}

The current generation of agent benchmarks shares a recipe: draw tasks from
economically valuable work, then score whether an agent completes them.
SWE-bench~\citep{jimenez2024swebench} and
SWE-Lancer~\citep{miserendino2025swelancer} score resolved GitHub issues and
paid freelance tickets; GAIA~\citep{mialon2024gaia} and
BrowseComp~\citep{wei2025browsecomp} score tool-use and browsing chains;
WebArena and its visual successor~\citep{zhou2024webarena,koh2024visualwebarena}
and OSWorld~\citep{xie2024osworld} score web and desktop workflows;
AgentBench~\citep{liu2024agentbench} aggregates across environments;
$\tau$-bench~\citep{yao2024taubench} adds a simulated user;
MLE-bench~\citep{chan2024mlebench} and PaperBench~\citep{starace2025paperbench}
score end-to-end research engineering. Occupation-scale suites such as
GDPval~\citep{patwardhan2025gdpval}, the Remote Labor
Index~\citep{mazeika2025rli}, APEX~\citep{vidgen2025apex}, and
ProfBench~\citep{wang2025profbench} extend the recipe toward whole professions.
These benchmarks reward usable artifacts produced in a stable, cooperative
environment. UserBench~\citep{qian2025userbench} documents one blind spot, agents that
guess rather than elicit what an underspecified goal leaves out (full
alignment with user intent on 20\% of tasks), and PRBench~\citep{akyurek2025prbench}
another, conclusions reached through inaccurate judgments and opaque,
incomplete reasoning on open-ended professional tasks. MA-Bench
targets those blind spots in planning, where the artifact is neither code nor a
booking but a staff product whose value lies in its defensibility against doctrine requirements.

\subsection{Evaluating planning}

Planning has long provided a particularly sharp lens on the gap between fluent generation
and grounded reasoning. PlanBench~\citep{valmeekam2023planbench} showed that
planning claims drawn from common-sense tasks do not survive classical
planning domains with automated plan validation (GPT-4 solves 34\% of
Blocksworld plan-generation instances);
follow-up work argues that chain-of-thought does not close the
gap~\citep{stechly2024chainofthoughtlessness} and that LLMs are better used as
components within verifier loops than as sole
planners~\citep{kambhampati2024llmmodulo}. Applied planning benchmarks reach
similar conclusions in richer settings: TravelPlanner~\citep{xie2024travelplanner}
reports single-digit success on multi-constraint itineraries;
NATURAL~PLAN~\citep{zheng2024naturalplan} shows that performance degrades as the number of constraints
increases, most sharply in trip and meeting planning. QuestBench~\citep{li2025questbench}
examines the complementary skill of asking the right clarifying question, building
on research into ambiguity and clarification
~\citep{min2020ambigqa,aliannejadi2020clariq,zhang2025clarify}.
Prompting scaffolds such as ReAct~\citep{yao2023react} and
Reflexion~\citep{shinn2023reflexion}, and search scaffolds such as Tree of
Thoughts~\citep{yao2023tot}, improve exploration and self-correction and have
been tested on question answering, text-based games, coding and puzzle-like
domains with verifiable end states. Mission analysis differs in kind: the ``plan'' is a structured
analytical product with no single correct answer and is graded on whether required
determinations (tasks, assumptions, CCIRs, risks) are present, sourced, and
doctrinally sound; this setting is not covered by any of the above benchmarks.
Reasoning-optimized models~\citep{openai2024o1,guo2025deepseekr1} raise scores
on verifiable domains, making a rubric-graded, non-verifiable domain a
natural next target.

\subsection{LLM-as-judge: capability and known failure modes}

Model-based grading is now standard where reference answers do not exist.
MT-Bench and Chatbot Arena established the feasibility of model-based grading and its correlation with human
preference~\citep{zheng2023mtbench,chiang2024arena}; G-Eval formalized
criterion-prompted judging~\citep{liu2023geval}; Prometheus trained open
evaluators using fine-grained scoring rubrics~\citep{kim2024prometheus};
RewardBench benchmarks the graders themselves~\citep{lambert2024rewardbench}.
The failure modes are equally well documented: position bias~\citep{wang2024fairevaluators}, length bias and its
correction~\citep{dubois2024lcalpaca}, and self-preference, in which evaluators
favor their own generations~\citep{panickssery2024selfpreference}.
MA-Bench's judge design responds directly to this literature. The judge is
selected by measured agreement with an expert panel, must cite the rubric item's embedded ground truth in every verdict, and
returns only failed items under an enforced schema; since unlisted items pass,
a malformed or truncated verdict would inflate positive items and penalize
negative ones, so such responses are rejected and retried, and the shortest
accepted verdict lists 13 items.
Section~\ref{sec:judge_sens} regrades the briefs under judges from two other providers and measures self-preference directly, finding none beyond 0.03.

\subsection{Rubric-graded evaluation of professional work}

A parallel thread replaces preference judgments with expert-authored,
criterion-level rubrics. HealthBench applies physician-authored criteria to
clinical dialogues~\citep{arora2025healthbench}; ProfBench extends
rubric grading across professional domains~\citep{wang2025profbench}; recent
work directly examines rubric generation and aggregation~\citep{gunjal2025rubrics,huang2025rubric,cook2024tick}, 
and rubric-style checklists now serve as reward signals for post-training~\citep{gunjal2025rubrics,viswanathan2025checklists}. 
MA-Bench adopts the same
philosophy of atomic, binary, self-contained criteria and adds three
properties not previously combined in this literature: every criterion carries a
doctrinal citation (JP~5-0~\citep{jp50}), a grounding tier that distinguishes 
extraction, synthesis, and inference, and a severity taxonomy in which
critical omissions and doctrinally detrimental behaviors are scored
asymmetrically. Grading targets what a brief determines, not the prose: fluency earns
nothing.

\subsection{Language models in military decision-making}

Interest in LLMs for military planning is no longer hypothetical: the Department of
Defense has moved from studying generative AI to piloting it across
operational and administrative applications~\citep{cdao2024tflima}, while how to evaluate them
remains an open question. Research to date has focused on behavior rather than
competence: escalation dynamics in simulated crises~\citep{rivera2024escalation},
behavioral divergence between expert humans and models in
wargames~\citep{lamparth2024wargames}, multi-agent historical
simulation~\citep{hua2024waragent}, and RAND's assessment of the limits of AI, including large
language models, for wargaming~\citep{geist2024limits}. COA-GPT~\citep{goecks2024coagpt}
conditions course-of-action generation on doctrine excerpts but evaluates the
results by simulated engagement outcomes, not by doctrinal correctness. Doctrinal publications such as JP~5-0, which specifies 
mission analysis~\citep{jp50}, JP~3-0~\citep{jp30}, and the CJCS manual
series~\citep{cjcsm3105}, provide the codified standard against which MA-Bench grades. 
Human-factors work on decision-support
trust~\citep{steyvers2024three,eigner2024determinants,cabitza2025xai} motivates
the severity-weighted, veto-carrying score MA-Bench reports to commanders.

\subsection{Benchmark validity}

Finally, MA-Bench's design responds to critiques of static-benchmark validity: contamination~\citep{sainz2023contamination}, the limits of static
knowledge-style suites~\citep{hendrycks2021mmlu}, and the gap between
leaderboard scores and deployed behavior~\citep{liang2023helm}. Scenarios are
synthetic and expert-validated, so no packet or rubric exists on the public
internet; withheld answer documents are quarantined from the system under
test by construction; and the protocol grades the brief alone, so any system
that produces one can be scored against the same standard.

\section{Corpus}
\label{app:corpus}

\begin{figure}[h]
\centering
\includegraphics[width=\textwidth]{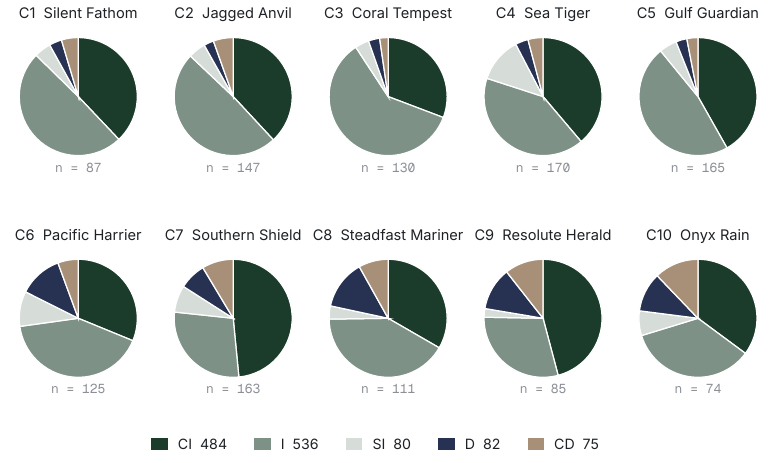}
\caption{\textbf{Severity composition of each scenario's rubric.} Wedges from
twelve o'clock clockwise: critically important, important, slightly important,
detrimental, critically detrimental. The authoring guideline kept every
category above 5\% of the corpus and every scenario's smallest category above
2\%, so the scoring behavior is not driven by any one scenario's mix.}
\label{fig:severity}
\end{figure}

Table~\ref{tab:corpus} lists the ten scenarios with the size of each packet and
rubric, and Figure~\ref{fig:severity} shows the severity mix of each rubric.

\begin{table}[h]
\centering
\footnotesize
\setlength{\tabcolsep}{6pt}
\fittable{%
\begin{tabular}{@{}l l r r r@{}}
\toprule
\colhead{Case} & \colhead{Operation} & \colhead{Documents} & \colhead{Packet chars} & \colhead{Rubric items} \\
\midrule
C1 & Silent Fathom & 12 & 134,337 & 87 \\
C2 & Jagged Anvil & 11 & 70,196 & 147 \\
C3 & Coral Tempest & 11 & 35,649 & 130 \\
C4 & Sea Tiger & 12 & 181,678 & 170 \\
C5 & Gulf Guardian & 12 & 80,341 & 165 \\
C6 & Pacific Harrier & 10 & 229,187 & 125 \\
C7 & Southern Shield & 12 & 242,658 & 163 \\
C8 & Steadfast Mariner & 11 & 156,729 & 111 \\
C9 & Resolute Herald & 11 & 151,265 & 85 \\
C10 & Onyx Rain & 10 & 93,897 & 74 \\
\midrule
\multicolumn{2}{@{}l}{Total} & & & 1,257 \\
\bottomrule
\end{tabular}}
\caption{\textbf{Scenarios.} Packet size counts the documents visible to the
system under test. Target distribution across the design axes: plan type
(contingency 60\%, crisis 25\%, campaign 15\%), domain (multi-domain 75\%,
single-domain 25\%), stakeholder mix (U.S.-only 25\%, bilateral 30\%,
multinational 35\%, whole-of-government 10\%), time pressure (deliberate
60\%, compressed 30\%, crisis 10\%), and operational-environment complexity
(permissive 25\%, contested 50\%, highly contested 25\%). Rubric totals by severity are CI 484, I 536, SI 80, D 82
and CD 75, and by tier 400, 406 and 451 for Tiers 1 to 3.}
\label{tab:corpus}
\end{table}

\FloatBarrier
\section{Additional rubric items and failure examples}
\label{app:cards}

A Tier~1 item, stated outright in the packet, and a critically detrimental
item, penalized when the brief commits the move it names, complement the
Tier~3 item shown in Section~\ref{sec:rubrics}.

\begin{rubriccard}{C5:GG-S1-02v1}{\meta{Tier 1}\metasep\meta{CI}\metasep\meta{Step S1}}
\small
\rubline{Question}{Does the response identify that the President declared a
No-Fly Zone over the Cuban coast and warned further killings would be met
forcefully?}
\rubline{Pass}{States the declared NFZ and the forceful-response warning.}
\rubline{Fail}{Omits the NFZ declaration or treats it as merely proposed.}
\rubline{Source}{Strategic Context; EXORD para.~3.}
\end{rubriccard}

\begin{rubriccard}{C2:R102}{\meta{Tier 2}\metasep\meta{CD}\metasep\meta{Step S6}}
\small
\rubline{Question}{Does the analysis treat the Iranian proxy attribution as
confirmed fact rather than a moderate-confidence assessment?}
\rubline{Basis}{The intelligence summary states the attribution is a
cross-source assessment, not a finding. A brief that hardens it into fact is
penalized.}
\end{rubriccard}

A third failure from the same Claude Opus 5 brief as the two in
Section~\ref{sec:failures}, on the commander's critical information
requirements step, which scores lowest of the twelve. The information
requirement exists but is not tied to the decision it should inform.

\begin{rubriccard}{C5:GG-S5-06}{\meta{Tier 3}\metasep\meta{I}\metasep\meta{Step S5}}
\small
\rubline{Item}{Does the response derive a friendly force information
requirement on branch readiness (ashore and regime change) and Nimitz carrier
strike group availability, tied to the recommendation to the National Command
Authority?}
\rubline{Brief}{FFIR 2 asks whether the Nimitz strike group and the amphibious
ready group will be available.}
\rubline{Judge}{``FFIR 2 addresses Nimitz and ARG availability but not
readiness of the ashore and regime-change branches or the NCA-recommendation
decision.''}
\end{rubriccard}

The Tier~1 counterpart of the card in Section~\ref{sec:failures}, from the
same brief, traced from the packet passage the item tests to the judge's
verdict, shows what the judge sees and what it decides on. The item belongs to the known-facts step. The packet states the
fact in one table. The brief, the first of Claude Opus 5's three on Gulf
Guardian, cites that paragraph three times, never reproduces it, and its
known-facts section does not mention partners at all. Another of the model's
three briefs on this scenario names all seven governments, so the item is
answerable within the word cap.

\begin{rubriccard}{C5:GG-S10-06v1, full trace}{\meta{Tier 1}\metasep\meta{I}\metasep\meta{Step S10}}
\small
\rubline{Packet}{Document 04, Coalition Order of Battle, paragraph 8, heading
and the four rows of its table: ``COALITION PARTNERS AND INTERNATIONAL
ORGANIZATIONS. Terms, consent, and capacity remain to be arranged.'' United
Kingdom: Caribbean patrol presence. Jamaica, Dominican Republic, Colombia,
Mexico: patrol, coordination; wary of onward migration. Trinidad and Tobago,
The Bahamas: willing to receive Cubans until conditions on the island change.
UN High Commissioner for Refugees (UNHCR): supervision of reception and
processing.}
\rubline{Item}{Does the response state the partner offers (United Kingdom,
Jamaica, Dominican Republic, Colombia, Mexico), the host nations (Trinidad and
Tobago, the Bahamas) and UNHCR?}
\rubline{Pass}{Names the partners, the host nations willing to receive Cubans,
and UNHCR.}
\rubline{Fail}{Refers to generic ``allies'' without naming partners or host
nations.}
\rubline{Brief}{\emph{Higher headquarters' intent:} ``act with willing
regional partners and sustain legitimacy of the civilian-protection case.''}
\rubline{Brief}{\emph{Tasks:} ``Negotiate partner terms, consent, and
capacity, and UNHCR supervision arrangements (Docs 04 para 8; 12).''}
\rubline{Brief}{\emph{CCIRs:} ``FFIR 4. Status of partner consent, capacity,
and UNHCR terms (Docs 04 para 8; 12). Supports burden-sharing decision.''}
\rubline{Brief}{\emph{Assumptions:} ``Willing regional partners and the UN
will provide burden-sharing reception once terms are agreed.''}
\rubline{Brief}{\emph{Known facts:} no mention of partners, host nations or
UNHCR.}
\rubline{Judge}{Fail. ``The brief refers generically to willing regional
partners and partner facilities but does not name the UK, Jamaica, Dominican
Republic, Colombia, Mexico, Trinidad and Tobago, and the Bahamas. The pass
condition requires the named partners, named host nations, and UNHCR.''}
\rubline{Score}{The item's weight of 5 is forgone. No penalty is added: only
the negative items penalize, and this brief committed none of the moves this
item is adjacent to.}
\end{rubriccard}

The brief points at the paragraph that holds the answer three times and never
reproduces it. A reader checking citations would pass this brief. The rubric
does not, and neither would a staff that had to call the embassies.

\FloatBarrier
\section{Scoring details}
\label{app:scoring}

A perfect brief scores 1 because the denominator is the total positive
weight, and wrong moves are charged in the same units as omissions. With items
weighted $+9$, $+5$, $+2$ and $-9$, passing the first two and avoiding the
mistake scores 0.875, passing all three but committing it scores 0.44, and
committing it with nothing passed scores $-0.56$. The weights do not decide the
ranking: weighted score and plain items-correct order the twenty systems of
Section~\ref{sec:results} the same way apart from swaps between adjacent models (Spearman 0.99).

When a report requires a bounded scale (a 0--100 score or letter bands),
clamping or an affine transformation is applied at reporting time only; the
unclamped score and item-level verdicts are always retained, so any mapping
can be recomputed. Counts of D and CD items are also reported directly, since
a single critically detrimental move can matter more to a commander than its
numerical weight conveys.

\FloatBarrier
\section{Main results table}
\label{app:main}

Table~\ref{tab:main} gives the numbers behind Figure~\ref{fig:main}: the
weighted score with its interval, the same score under the Gemini cross-judge,
and the unweighted measures.

\begin{table}[h]
\centering
\footnotesize
\setlength{\tabcolsep}{3.5pt}
\fittable{%
\begin{tabular}{@{}l r l r r r r@{}}
\toprule
\colhead{Model} & \colhead{Briefs} & \colhead{Score (95\% CI)} & \colhead{Gemini judge} & \colhead{Correct \%} & \colhead{Passed \%} & \colhead{Wrong moves (CD)} \\
\midrule
Muse Spark 1.3 & 30 & \textbf{0.574} (0.504, 0.646) & \textbf{0.597} & \textbf{61.8} & \textbf{59.1} & 3.5 (1.8) \\
Grok 4.6 & 30 & 0.540 (0.458, 0.625) & 0.545 & 58.3 & 55.3 & 4.1 (2.0) \\
Claude Opus 5 & 30 & 0.514 (0.427, 0.600) & 0.549 & 56.2 & 52.9 & 4.0 (2.1) \\
GLM-5.3 $\circ$ & 30 & 0.505 (0.431, 0.579) & 0.522 & 55.6 & 51.6 & 3.4 (1.8) \\
GPT-6 Astra & 50 & 0.485 (0.411, 0.556) & 0.532 & 53.4 & 48.9 & \textbf{3.1} (1.4) \\
Kimi K3 $\circ$ & 30 & 0.481 (0.414, 0.550) & 0.541 & 53.3 & 50.7 & 5.0 (2.7) \\
GPT-5.6 Sol & 30 & 0.460 (0.376, 0.541) & 0.496 & 51.2 & 47.4 & 4.1 (2.1) \\
GPT-5.6 Luna & 30 & 0.452 (0.362, 0.539) & 0.487 & 50.9 & 47.3 & 4.5 (2.4) \\
Nemotron 3 Ultra $\circ$ & 30 & 0.434 (0.339, 0.518) & 0.441 & 48.9 & 44.3 & 3.9 (2.0) \\
MiniMax M3 $\circ$ & 30 & 0.430 (0.343, 0.508) & 0.444 & 49.2 & 44.8 & 3.9 (1.8) \\
Claude Fable 5.1 $\dagger$ & 50 & 0.428 (0.282, 0.558) & 0.471 & 49.2 & 46.0 & 4.9 (2.5) \\
Qwen3.8-Max & 30 & 0.410 (0.326, 0.494) & 0.493 & 47.4 & 43.6 & 4.8 (2.7) \\
Claude Sonnet 5 $\diamond$ & 30 & 0.393 (0.326, 0.463) & 0.462 & 45.3 & 41.0 & 4.3 (2.1) \\
Gemini 3.8 Flash $\diamond$ & 30 & 0.371 (0.286, 0.453) & 0.351 & 43.6 & 38.6 & 4.0 (1.9) \\
GPT-5.4 $\diamond$ & 30 & 0.370 (0.284, 0.455) & 0.402 & 43.6 & 38.9 & 4.6 (2.2) \\
DeepSeek-V4-Pro $\circ$ & 30 & 0.359 (0.272, 0.444) & 0.393 & 42.7 & 37.4 & 4.3 (2.0) \\
Gemini 3.1 Pro & 50 & 0.340 (0.256, 0.425) & 0.365 & 40.9 & 36.2 & 5.0 (2.4) \\
Mistral Medium 3.5 & 30 & 0.338 (0.240, 0.429) & 0.389 & 41.3 & 36.3 & 4.7 (2.3) \\
DeepSeek-V4-Flash $\diamond$ & 30 & 0.307 (0.220, 0.396) & 0.334 & 38.0 & 32.8 & 5.0 (2.3) \\
Gemma 4 31B $\circ$ & 30 & 0.302 (0.229, 0.380) & 0.341 & 37.8 & 32.2 & 4.5 (2.1) \\
\bottomrule
\end{tabular}}
\caption{\textbf{Benchmark condition, production judge}, 3,000-word cap and
medium reasoning effort, sorted by weighted score. Score is the unclamped
severity-weighted score of Section~\ref{sec:scoring} with a 95\% bootstrap
interval over scenarios. Gemini judge is the same score under the Gemini 3.1
Pro cross-judge. Correct is the share of all items graded correctly, Passed
the share of positive items satisfied, and Wrong moves the mean number of
negative items committed per brief with the critically detrimental subset in
parentheses. Best value per column in bold. $\circ$ open weights.
$\diamond$ smaller member of a family, run as the capability ladder.
$\dagger$ includes nine content-filter refusals graded as delivered.}
\label{tab:main}
\end{table}

\FloatBarrier
\section{Generation statistics}
\label{app:gen}

Table~\ref{tab:gen} records how long each system wrote before truncation, how
many of its briefs hit the 3,000-word cap, and how many ended in a provider
content filter.

\begin{table}[h]
\centering
\footnotesize
\setlength{\tabcolsep}{5pt}
\fittable{%
\begin{tabular}{@{}l l r l r r@{}}
\toprule
\colhead{Model} & \colhead{Condition} & \colhead{Briefs} & \colhead{Words, mean (min to max)} & \colhead{Truncated} & \colhead{Refusals} \\
\midrule
Claude Opus 5 & cap, medium & 30 & 3,654 (2,846 to 7,069) & 27 & 0 \\
Claude Fable 5.1 & cap, medium & 50 & 2,811 (101 to 3,798) & 39 & 9 \\
GPT-6 Astra & cap, medium & 50 & 2,333 (2,146 to 2,504) & 0 & 0 \\
GPT-5.6 Sol & cap, default & 30 & 2,392 (2,111 to 2,826) & 0 & 0 \\
GPT-5.6 Luna & cap, default & 30 & 3,034 (2,503 to 3,517) & 17 & 0 \\

Gemini 3.1 Pro & cap, medium & 50 & 2,090 (1,638 to 2,569) & 0 & 0 \\
Qwen3.8-Max & cap, medium & 30 & 2,416 (2,038 to 2,939) & 0 & 0 \\
GLM-5.3 & cap, medium & 30 & 3,155 (2,617 to 3,778) & 21 & 0 \\
DeepSeek-V4-Pro & cap, default & 30 & 4,701 (2,255 to 8,141) & 23 & 0 \\
Gemma 4 31B & cap, default & 30 & 1,663 (1,458 to 2,089) & 0 & 0 \\
Grok 4.6 & cap, default & 30 & 3,341 (2,687 to 3,967) & 25 & 0 \\
Kimi K3 & cap, default & 30 & 2,819 (2,247 to 3,576) & 10 & 0 \\
MiniMax M3 & cap, default & 30 & 3,356 (1,997 to 4,464) & 23 & 0 \\
Mistral Medium 3.5 & cap, default & 30 & 3,563 (1,716 to 6,995) & 18 & 0 \\
Nemotron 3 Ultra & cap, default & 30 & 4,569 (2,781 to 7,610) & 29 & 0 \\
Muse Spark 1.3 & cap, medium & 30 & 3,315 (2,703 to 4,429) & 23 & 0 \\
\addlinespace
Claude Sonnet 5 & cap, medium & 30 & 2,034 (1,482 to 2,668) & 0 & 0 \\
GPT-5.4 & cap, medium & 30 & 2,111 (1,794 to 2,296) & 0 & 0 \\
Gemini 3.8 Flash & cap, medium & 30 & 5,239 (4,326 to 6,940) & 30 & 0 \\
DeepSeek-V4-Flash & cap, default & 30 & 2,435 (1,444 to 3,898) & 4 & 0 \\
\addlinespace
GPT-6 Astra & cap, low & 30 & 2,260 (2,100 to 2,408) & 0 & 0 \\
GPT-6 Astra & cap, high & 30 & 2,375 (2,209 to 2,523) & 0 & 0 \\
GPT-5.6 Sol & cap, low & 30 & 2,344 (2,125 to 2,746) & 0 & 0 \\
GPT-5.6 Sol & cap, high & 30 & 2,399 (2,070 to 2,722) & 0 & 0 \\
Claude Sonnet 5 & cap, low & 30 & 1,584 (1,149 to 1,955) & 0 & 0 \\
Claude Sonnet 5 & cap, high & 30 & 2,321 (2,019 to 2,712) & 0 & 0 \\
GPT-5.4 & cap, low & 30 & 2,065 (1,847 to 2,553) & 0 & 0 \\
GPT-5.4 & cap, high & 30 & 2,246 (1,880 to 2,727) & 0 & 0 \\
Gemini 3.8 Flash & cap, low & 30 & 4,922 (3,518 to 6,458) & 30 & 0 \\
Gemini 3.8 Flash & cap, high & 30 & 5,453 (4,453 to 6,649) & 30 & 0 \\
GPT-6 Astra & uncapped, medium & 30 & 7,084 (5,794 to 8,728) & 0 & 0 \\
GPT-5.4 & uncapped & 30 & 5,898 (5,087 to 6,858) & 0 & 0 \\
\bottomrule
\end{tabular}}
\caption{\textbf{Brief lengths.} Word counts are of the raw output before
truncation. Truncated counts briefs longer than 3,000 words that were cut at
the cap before grading. Default marks runs at the provider's default reasoning
effort, because the interface does not expose the parameter or it was left
unset. Refusals are outputs that ended in a
provider content filter.}
\label{tab:gen}
\end{table}

\FloatBarrier
\section{Decomposition by mission analysis step}
\label{app:steps}

\begin{figure}[h]
\centering
\includegraphics[width=\textwidth]{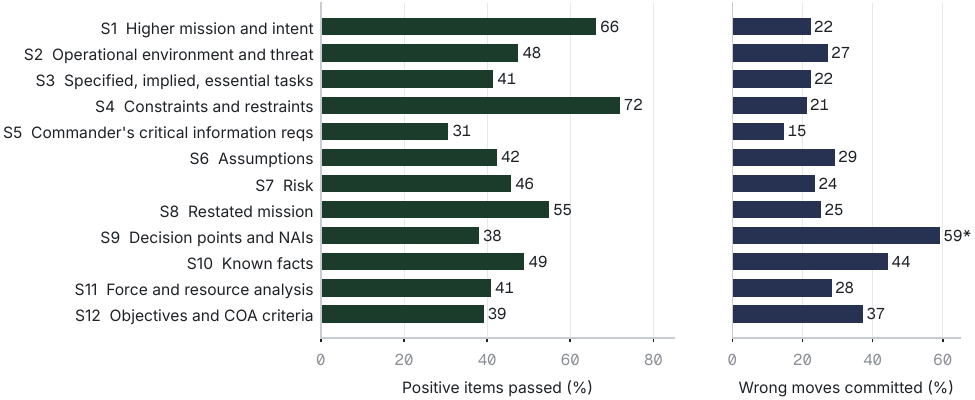}
\caption{\textbf{Coverage and wrong moves by mission analysis step},
benchmark condition, all models pooled. Left, share of positive items passed.
Right, share of negative items on which the flagged mistake was committed
(asterisk: fewer than 100 graded negative items).}
\label{fig:steps}
\end{figure}

Figure~\ref{fig:steps} and Table~\ref{tab:steps} decompose the pooled
benchmark-condition results by mission analysis step, and
Table~\ref{tab:tiersev} by tier and severity.

\begin{table}[h]
\centering
\footnotesize
\setlength{\tabcolsep}{5pt}
\fittable{%
\begin{tabular}{@{}l l r r r r@{}}
\toprule
\colhead{Step} & & \colhead{Positive graded} & \colhead{Pass rate} & \colhead{Negative graded} & \colhead{Committed} \\
\midrule
S1 & Higher mission and intent & 5,832 & 66.2\% & 810 & 22.3\% \\
S2 & Operational environment and threat & 9,828 & 47.5\% & 972 & 27.3\% \\
S3 & Specified, implied and essential tasks & 8,046 & 41.5\% & 648 & 22.4\% \\
S4 & Constraints and restraints & 3,456 & 71.9\% & 1,674 & 21.2\% \\
S5 & Commander's critical information requirements & 7,182 & 30.7\% & 108 & 14.8\% \\
S6 & Assumptions & 4,752 & 42.3\% & 810 & 29.3\% \\
S7 & Risk & 3,348 & 45.8\% & 594 & 23.6\% \\
S8 & Restated mission & 1,674 & 54.9\% & 810 & 25.3\% \\
S9 & Decision points and named areas of interest & 1,566 & 38.0\% & 54 & 59.3\% \\
S10 & Known facts & 4,374 & 49.0\% & 702 & 44.4\% \\
S11 & Force and resource analysis & 5,778 & 40.9\% & 864 & 28.4\% \\
S12 & Military objectives and COA evaluation criteria & 3,564 & 39.1\% & 432 & 37.3\% \\
\bottomrule
\end{tabular}}
\caption{\textbf{Coverage and wrong moves by step}, benchmark condition,
all models pooled, production judge. Graded counts are item verdicts.
Committed is the share of negative items on which the judge found the
mistake present.}
\label{tab:steps}
\end{table}

\begin{table}[h]
\centering
\footnotesize
\setlength{\tabcolsep}{6pt}
\fittable{%
\begin{tabular}{@{}l r r r r@{}}
\toprule
\colhead{Tier} & \colhead{Positive graded} & \colhead{Pass rate} & \colhead{Negative graded} & \colhead{Committed} \\
\midrule
Tier 1, stated & 21,222 & 58.1\% & 378 & 34.7\% \\
Tier 2, synthesized & 15,498 & 49.0\% & 6,426 & 28.2\% \\
Tier 3, inferred & 22,680 & 33.4\% & 1,674 & 21.0\% \\
\midrule
\colhead{Severity} & & & & \\
\midrule
CI & 26,136 & 56.9\% & & \\
I & 28,944 & 37.5\% & & \\
SI & 4,320 & 41.5\% & & \\
D & & & 4,428 & 25.9\% \\
CD & & & 4,050 & 28.3\% \\
\bottomrule
\end{tabular}}
\caption{\textbf{Coverage and wrong moves by tier and severity}, benchmark
condition, all models pooled, production judge.}
\label{tab:tiersev}
\end{table}

\FloatBarrier
\section{Failure modes by model}
\label{app:failures}

\begin{figure}[h]
\centering
\includegraphics[width=\textwidth]{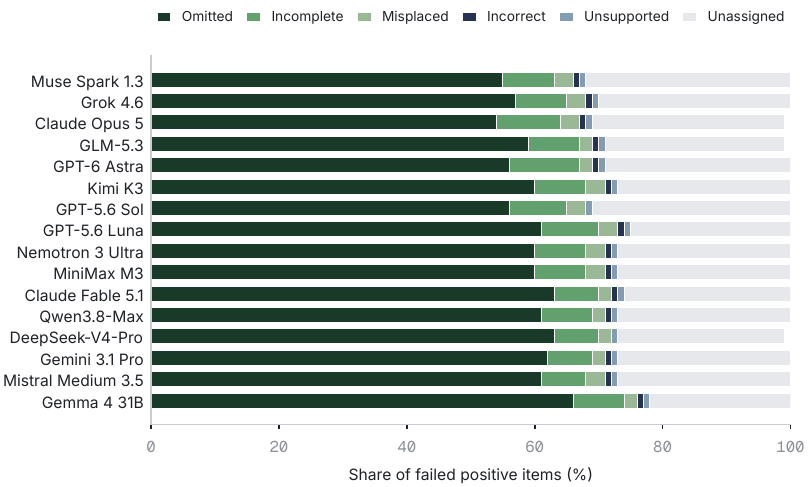}
\caption{\textbf{Failure composition by model.} The omission share varies by
only twelve points across sixteen models.}
\label{fig:failures}
\end{figure}

Table~\ref{tab:taxtier} classifies the failed positive items by tier and
Table~\ref{tab:failures} by model, using the keyword rules of
Section~\ref{sec:failures}; Figure~\ref{fig:failures} draws the per-model
composition.

\begin{table}[h]
\centering
\footnotesize
\setlength{\tabcolsep}{4.5pt}
\fittable{%
\begin{tabular}{@{}l r r r r r r r r@{}}
\toprule
\colhead{Tier} & \colhead{Failed} & \colhead{Omitted} & \colhead{Incomplete} & \colhead{Misplaced} & \colhead{Incorrect} & \colhead{Unsupported} & \colhead{Unassigned} & \colhead{Wrong moves} \\
\midrule
Tier 1, stated & 41.9\% & 66.7\% & 6.7\% & 2.3\% & 0.4\% & 0.5\% & 23.4\% & 34.7\% \\
Tier 2, synthesized & 51.0\% & 54.0\% & 8.7\% & 3.1\% & 1.4\% & 1.7\% & 31.0\% & 28.2\% \\
Tier 3, inferred & 66.6\% & 59.5\% & 8.7\% & 2.2\% & 0.4\% & 0.9\% & 28.2\% & 21.0\% \\
\midrule
All & 53.7\% & 60.2\% & 8.2\% & 2.5\% & 0.7\% & 1.0\% & 27.6\% & 27.1\% \\
\bottomrule
\end{tabular}}
\caption{\textbf{How positive items fail, by tier}, benchmark condition,
production judge. Failed is the share of positive items failed (21,222, 15,498 and 22,680 graded by tier). The next six columns classify the failed
items from the judge's evidence and rationale by keyword rules and sum to
100\%. Wrong moves is the share of negative items on which the flagged mistake
was committed (378, 6,426 and 1,674 graded by tier).}
\label{tab:taxtier}
\end{table}

\begin{table}[h]
\centering
\footnotesize
\setlength{\tabcolsep}{5pt}
\fittable{%
\begin{tabular}{@{}l r r r r r r r@{}}
\toprule
\colhead{Model} & \colhead{Failed items} & \colhead{Omitted} & \colhead{Incomplete} & \colhead{Misplaced} & \colhead{Incorrect} & \colhead{Unsupported} & \colhead{Unassigned} \\
\midrule
Muse Spark 1.3 & 1,333 & 55\% & 8\% & 3\% & 1\% & 1\% & 32\% \\
Grok 4.6 & 1,441 & 57\% & 8\% & 3\% & 1\% & 1\% & 30\% \\
Claude Opus 5 & 1,516 & 54\% & 10\% & 3\% & 1\% & 1\% & 30\% \\
GLM-5.3 & 1,580 & 59\% & 8\% & 2\% & 1\% & 1\% & 28\% \\
Kimi K3 & 1,599 & 60\% & 8\% & 3\% & 1\% & 1\% & 27\% \\
GPT-5.6 Sol & 1,705 & 56\% & 9\% & 3\% & 0\% & 1\% & 31\% \\
GPT-5.6 Luna & 1,707 & 61\% & 9\% & 3\% & 1\% & 1\% & 26\% \\
MiniMax M3 & 1,814 & 60\% & 8\% & 3\% & 1\% & 1\% & 27\% \\
Nemotron 3 Ultra & 1,818 & 60\% & 8\% & 3\% & 1\% & 1\% & 27\% \\
Qwen3.8-Max & 1,831 & 61\% & 8\% & 2\% & 1\% & 1\% & 27\% \\
DeepSeek-V4-Pro & 2,031 & 63\% & 7\% & 2\% & 0\% & 1\% & 26\% \\
Mistral Medium 3.5 & 2,081 & 61\% & 7\% & 3\% & 1\% & 1\% & 27\% \\
Gemma 4 31B & 2,218 & 66\% & 8\% & 2\% & 1\% & 1\% & 23\% \\
GPT-6 Astra & 2,770 & 56\% & 11\% & 2\% & 1\% & 1\% & 30\% \\
Claude Fable 5.1 & 2,978 & 63\% & 7\% & 2\% & 1\% & 1\% & 27\% \\
Gemini 3.1 Pro & 3,460 & 62\% & 7\% & 2\% & 1\% & 1\% & 27\% \\
\midrule
All & 31,882 & 60.2\% & 8.2\% & 2.5\% & 0.7\% & 1.0\% & 27.6\% \\
\bottomrule
\end{tabular}}
\caption{\textbf{Failure modes of positive items}, benchmark condition,
production judge. Buckets are assigned from the judge's rationale by keyword
rules. Omitted means the determination is absent, incomplete that it is
present but missing a required element, misplaced that it appears under the
wrong step, incorrect that it contradicts the source, unsupported that it is
asserted without the source's basis. Unassigned is the residue the rules did
not classify.}
\label{tab:failures}
\end{table}

\FloatBarrier
\section{Use of large language models}
\label{app:llm}

Language models are both the object of study and part of the instrument. The
graded briefs are model outputs by design, and the judge is a language model
whose agreement with human experts is measured in Section~\ref{sec:irr}. An
LLM-based coding assistant was also used to help write analysis and plotting
code and to draft and edit prose. The authors designed the benchmark and the
experiments, ran them, and verified every number and claim in the paper.

\FloatBarrier
\section{Length and effort ablations}
\label{app:ablation}

\begin{figure}[h]
\centering
\includegraphics[width=\textwidth]{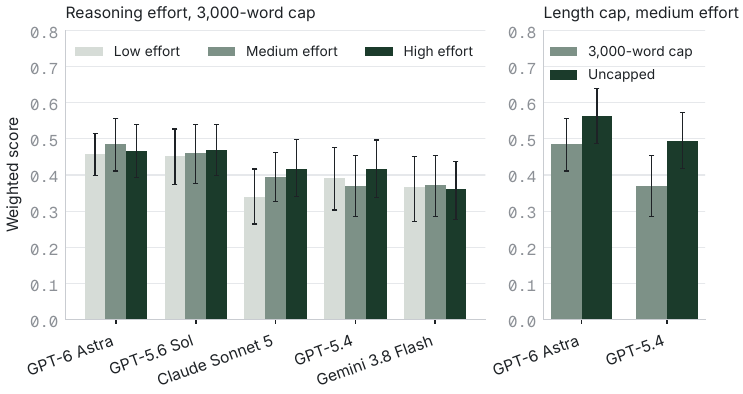}
\caption{\textbf{Reasoning effort and the length cap.} Left: weighted score
with 95\% intervals at low, medium and high reasoning effort under the
3,000-word cap for five models. Right: the same score capped and uncapped for
GPT-6 Astra and GPT-5.4.}
\label{fig:length}
\end{figure}

Removing the 3,000-word cap raises GPT-6 Astra from 0.485 to 0.562, a paired
difference of $+0.078$ (95\% interval $+0.024$ to $+0.125$) at three times the
length, and GPT-5.4 from 0.370 to 0.493 ($+0.123$) (Figure~\ref{fig:length},
Table~\ref{tab:ablation}). Models therefore hold more rubric-relevant content
than a staff-length brief can carry, and the benchmark condition measures
prioritization as well as knowledge. Length alone does not buy coverage:
Gemini 3.8 Flash writes the longest capped briefs, 5,200 words before
truncation, and scores 0.371, while Qwen3.8-Max never reaches the cap and
scores 0.410. Reasoning effort was swept from low to high for five models.
The largest movement is Claude Sonnet 5, which loses 0.056 at low effort and
gains 0.023 at high effort relative to medium; GPT-5.4 gains 0.045 at high
effort; GPT-6 Astra, GPT-5.6 Sol and Gemini 3.8 Flash move by less than 0.03
in either direction, and Astra is slightly worse at high effort than at
medium. Across the sweep the reasoning tokens reported by the providers rise
by one to two orders of magnitude from low to high (GPT-6 Astra 24 to 3,533,
GPT-5.6 Sol 170 to 2,442, GPT-5.4 67 to 7,467, Gemini 3.8 Flash 9,742 at
high; the Anthropic interface does not report them) while scores move by at
most 0.08, less than a third of the 0.27 spread between models in the
benchmark condition. Within a family the stronger member wins in three of four cases,
Claude Opus 5 over Sonnet 5 by 0.12, GPT-6 Astra over GPT-5.4 by 0.115 and
DeepSeek-V4-Pro over V4-Flash by 0.05, with the Gemini pair tied.

\begin{table}[h]
\centering
\footnotesize
\setlength{\tabcolsep}{4pt}
\fittable{%
\begin{tabular}{@{}l l r l r l@{}}
\toprule
\colhead{Model} & \colhead{Condition} & \colhead{Words} & \colhead{Weighted score (95\% CI)} & \colhead{Gemini judge} & \colhead{Paired difference} \\
\midrule
GPT-6 Astra & cap, medium & 2,333 & 0.485 (0.411, 0.556) & 0.532 & reference \\
GPT-6 Astra & cap, low & 2,260 & 0.459 (0.398, 0.516) & 0.534 & $-0.026$ ($-0.059$, $+0.005$) \\
GPT-6 Astra & cap, high & 2,375 & 0.467 (0.394, 0.541) & 0.504 & $-0.018$ ($-0.031$, $-0.004$) \\
GPT-6 Astra & uncapped, medium & 7,084 & 0.562 (0.487, 0.641) & 0.639 & $+0.078$ ($+0.024$, $+0.125$) \\
\addlinespace
GPT-5.6 Sol & cap, medium & 2,392 & 0.460 (0.376, 0.541) & 0.496 & reference \\
GPT-5.6 Sol & cap, low & 2,344 & 0.451 (0.374, 0.527) & 0.504 & $-0.009$ ($-0.028$, $+0.011$) \\
GPT-5.6 Sol & cap, high & 2,399 & 0.469 (0.398, 0.539) & 0.498 & $+0.009$ ($-0.005$, $+0.026$) \\
\addlinespace
Claude Sonnet 5 & cap, medium & 2,034 & 0.393 (0.326, 0.463) & 0.462 & reference \\
Claude Sonnet 5 & cap, low & 1,584 & 0.338 (0.265, 0.417) & 0.433 & $-0.056$ ($-0.080$, $-0.031$) \\
Claude Sonnet 5 & cap, high & 2,321 & 0.417 (0.340, 0.498) & 0.489 & $+0.023$ ($-0.002$, $+0.048$) \\
\addlinespace
GPT-5.4 & cap, medium & 2,111 & 0.370 (0.284, 0.455) & 0.402 & reference \\
GPT-5.4 & cap, low & 2,065 & 0.391 (0.303, 0.476) & 0.420 & $+0.021$ ($-0.001$, $+0.044$) \\
GPT-5.4 & cap, high & 2,246 & 0.415 (0.337, 0.497) & 0.473 & $+0.045$ ($+0.002$, $+0.079$) \\
GPT-5.4 & uncapped, medium & 5,898 & 0.493 (0.418, 0.573) & 0.535 & $+0.123$ ($+0.076$, $+0.170$) \\
\addlinespace
Gemini 3.8 Flash & cap, medium & 5,239 & 0.371 (0.286, 0.453) & 0.351 & reference \\
Gemini 3.8 Flash & cap, low & 4,922 & 0.367 (0.272, 0.451) & 0.383 & $-0.004$ ($-0.029$, $+0.024$) \\
Gemini 3.8 Flash & cap, high & 5,453 & 0.362 (0.278, 0.438) & 0.365 & $-0.009$ ($-0.032$, $+0.016$) \\
\bottomrule
\end{tabular}}
\caption{\textbf{Length and effort ablations.} Words is the mean length of the
raw output before truncation. Differences are
paired by scenario against the medium-effort row of the same model with 95\%
bootstrap intervals; Gemini judge is the same score under the Gemini 3.1 Pro
cross-judge.}
\label{tab:ablation}
\end{table}

\begin{figure}[h]
\centering
\includegraphics[width=\textwidth]{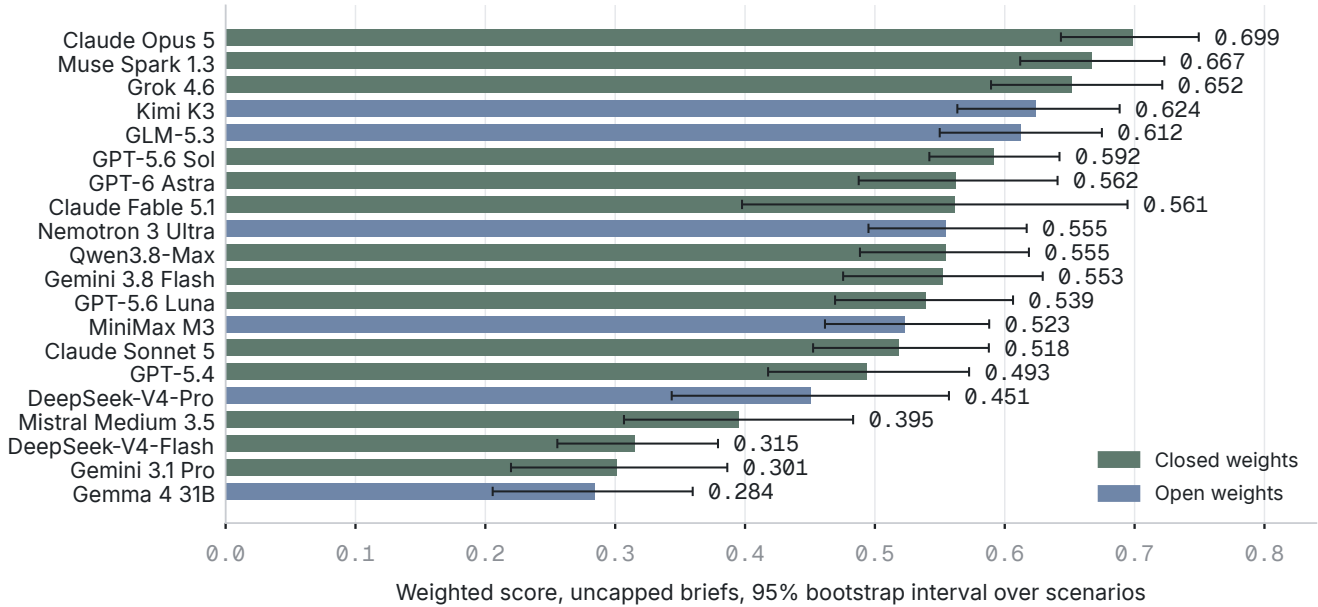}
\caption{\textbf{Weighted score by model without the length cap}, production
judge, three briefs per scenario, with 95\% bootstrap intervals over the ten
scenarios. The GPT-6 Astra and GPT-5.4 rows are the uncapped runs of
Table~\ref{tab:ablation}.}
\label{fig:uncapped}
\end{figure}

Every model was also run without the cap, with the benchmark prompt minus the
length limit and the model settings of Section~\ref{sec:setup}, three briefs
per scenario (Figure~\ref{fig:uncapped}). The ranking largely holds, with a
Spearman correlation of 0.95 against the benchmark condition. The average
model gains 0.098, eighteen of the twenty score higher, and Claude Opus 5
moves from third to first at 0.699. Gemini 3.1 Pro and Gemma 4 31B score
slightly lower, and both write under 1,800 words even without the cap.
Without the cap the score also tracks length far more closely, with a Spearman
correlation of 0.70 between a model's mean brief length and its score against
0.26 under the cap.

\FloatBarrier
\section{Scores under the alternate judges}
\label{app:judges}

\begin{figure}[h]
\centering
\includegraphics[width=0.6\textwidth]{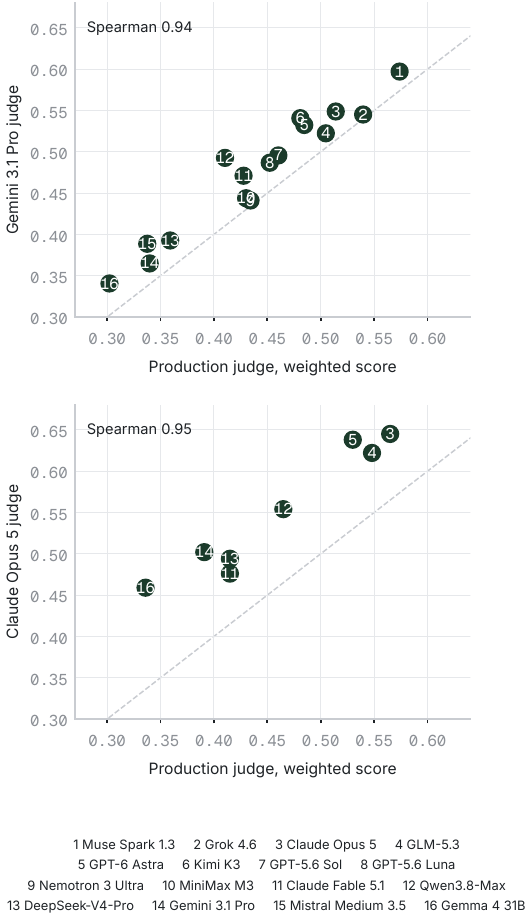}
\caption{\textbf{Model means under three judges.} Each point is one
benchmark-condition model. Both alternate judges sit above the identity line by
a roughly constant offset and preserve the order (Spearman 0.94 over the sixteen benchmark-condition
models for Gemini, 0.94 on the common set for Opus).}
\label{fig:judges}
\end{figure}

Table~\ref{tab:candidates} compares the candidate judges against the expert
panel, Table~\ref{tab:altjudges} gives each model's score under the three
judges on the common brief set, and Figure~\ref{fig:judges} plots the two
alternate judges against the production judge. Two robustness checks on the
expert sample: a repeat run of the frozen judge on the same 190 items moved
$\kappa$ by 0.004, and on the 137 items that the final rubric revision left
unchanged the judge's $\kappa$ is 0.51, so the result does not rest on the
revised items.

\begin{table}[h]
\centering
\footnotesize
\setlength{\tabcolsep}{4pt}
\fittable{%
\begin{tabular}{@{}l r r r@{}}
\toprule
\colhead{Candidate judge} & \colhead{Agreement} & \colhead{Cohen kappa} & \colhead{F1, fail class} \\
\midrule
gpt-5.6, frozen prompt & \textbf{78.0\%} & \textbf{0.540} & \textbf{0.822} \\
gpt-5.6, softened rules & 73.2\% & 0.459 & 0.771 \\
gpt-5.5, frozen prompt & 70.1\% & 0.405 & 0.741 \\
Gemini 3.1 Pro, frozen prompt & 72.6\% & 0.427 & 0.776 \\
Claude Opus 5, frozen prompt & 64.6\% & 0.339 & 0.663 \\
\bottomrule
\end{tabular}}
\caption{\textbf{Candidate judges against the expert majority} on the 164
majority-verdict items of the 190-item expert study. For reference, two
experts agree with each other at a mean pairwise kappa of 0.43, the ceiling
against which a judge is read. Best in bold.}
\label{tab:candidates}
\end{table}

\begin{table}[h]
\centering
\footnotesize
\setlength{\tabcolsep}{6pt}
\fittable{%
\begin{tabular}{@{}l r r r@{}}
\toprule
\colhead{Model} & \colhead{gpt-5.6 (production)} & \colhead{Gemini 3.1 Pro} & \colhead{Claude Opus 5} \\
\midrule
Claude Opus 5 & \textbf{0.565} & 0.591 & \textbf{0.645} \\
GLM-5.3 & 0.548 & 0.569 & 0.622 \\
GPT-6 Astra & 0.530 & \textbf{0.595} & 0.638 \\
GPT-5.6 Sol & 0.514 & 0.524 & 0.615 \\
GPT-5.6 Luna & 0.505 & 0.540 & 0.609 \\
Claude Fable 5.1 & 0.415 & 0.463 & 0.476 \\
Qwen3.8-Max & 0.465 & 0.544 & 0.554 \\
DeepSeek-V4-Pro & 0.415 & 0.437 & 0.494 \\
Gemini 3.1 Pro & 0.391 & 0.427 & 0.502 \\
Gemma 4 31B & 0.336 & 0.385 & 0.459 \\
\bottomrule
\end{tabular}}
\caption{\textbf{Weighted score under each judge on the common set} of 216
benchmark-condition briefs (six scenarios, ten models, every seed), same frozen
prompt, high reasoning effort. The alternate judges are more lenient than the
production judge by a roughly constant offset and preserve its ordering
(Spearman 0.92 and 0.94, Table~\ref{tab:judges}). Best per column in bold.}
\label{tab:altjudges}
\end{table}

\FloatBarrier
\clearpage
\section{Prompts}
\label{app:prompts}

The task prompt below is fixed across every model and condition. Only the
operation name varies with the case, and the final paragraph states the
length cap. The system under test receives this prompt followed by the
heading \texttt{\# SOURCE DOCUMENTS} and the packet.

\promptlisting{prompts/task_prompt.txt}

The judge receives the system prompt below, then a user message built from the
template that follows, into which the case header, the task prompt, the case's
rubric items as JSON (id, step, question, pass condition, fail condition,
source fact, artifact anchor, source, tier and severity) and the brief are
inserted. The judge never receives the packet.

\promptlisting{prompts/judge_system.txt}

\promptlisting{prompts/judge_user.txt}

\end{document}